\documentclass{article}
\usepackage{dingbat}
\usepackage{spconf,amsmath,graphicx}
\usepackage{amsfonts}
\usepackage{xcolor}
\usepackage{hyperref}
\usepackage{xspace}\usepackage[compact]{titlesec}         

\title{BOSS: A Benchmark for Human Belief Prediction in Object-context Scenarios}
\name{Jiafei Duan$^{1\ast}$ \qquad Samson Yu$^{2\ast}$ \qquad Nicholas Tan$^{3}$ \qquad Li Yi$^{4}$ \qquad Cheston Tan$^{2}$}

\address{\{duan\_jiafei, cheston-tan\}@i2r.a-star.edu.sg, samson\_yu@ihpc.a-star.edu.sg, \\nicholas\_tan\_yun\_yu@u.nus.edu, ericyi0124@gmail.com\\
$^{1}$Institute for Infocomm Research, A*STAR\\
$^{2}$Centre for Frontier AI Research, A*STAR\\
$^{3}$Department of Computer Science, National University of Singapore\\
$^{4}$Institute for Interdisciplinary Information Sciences, Tsinghua University\\
}

\begin{document}
\setlength\abovedisplayskip{0pt}
\setlength\belowdisplayskip{0pt}
%
\maketitle
\begin{abstract}
Humans with an average level of social cognition can infer the beliefs of others based solely on the nonverbal communication signals (e.g. gaze, gesture, pose and contextual information) exhibited during social interactions \cite{hinde1974biological,wellman2001meta,saxe2006uniquely}. This social cognitive ability to predict human beliefs and intentions is more important than ever for ensuring safe human-robot interaction and collaboration. This paper uses the combined knowledge of \textbf{Theory of Mind (ToM)} and \textbf{Object-Context Relations} to investigate methods for enhancing collaboration between humans and autonomous systems in environments where verbal communication is prohibited. We propose a novel and challenging multimodal video dataset for assessing the capability of artificial intelligence (AI) systems in predicting human belief states in an object-context scenario. The proposed dataset consists of precise labelling of human belief state ground-truth and multimodal inputs replicating all nonverbal communication inputs captured by human perception. We further evaluate our dataset with existing deep learning models and provide new insights into the effects of the various input modalities and object-context relations on the performance of the baseline models.

\end{abstract}

\section{Introduction}
\label{intro}
\begin{figure*}[ht]
    \centering
    \includegraphics[width=\linewidth]{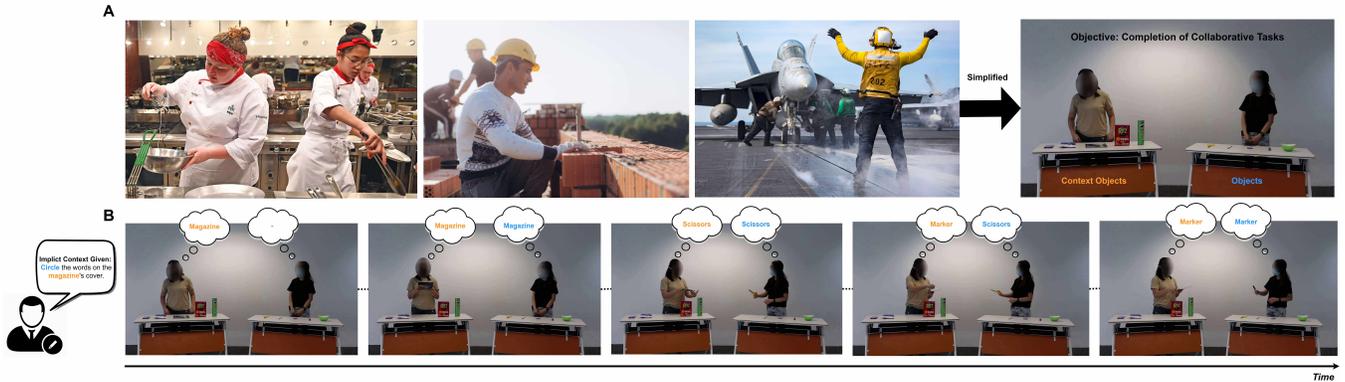}
    
    \caption{(A) Real-world examples of collaborative tasks that require inferring each other's beliefs via nonverbal communication. (B) An example of the instructions provided during data collection.}
\label{fig:0}
\end{figure*}

People sometimes have the illusion of having the ability to `read' someone's mind just by observing their actions and behaviours. In reality, we cannot read the minds of others, but we can create a mental model of their belief states based on the various nonverbal communication signals (e.g. gaze, gesture, pose and context) that we receive in a social interaction context. This social cognitive ability is especially vital in a number of settings, where verbal communication between people is limited, while a collaborative task is required to be accomplished (e.g. in a loud factory, a crowded location or even a noisy restaurant kitchen), as shown in Figure \ref{fig:0}A. With the advancement of embodied AI \cite{duan2021survey}, robotics\cite{argall2009survey} and computer vision \cite{voulodimos2018deep}, many of these settings will potentially have autonomous systems working closely with humans in a collaborative manner. Hence, research on this social cognitive ability to infer human belief state may become increasingly crucial for robotics/AI systems, but is under-studied in scene understanding, robot perception and computer vision.

In the study of cognitive science, researchers decompose this social cognitive ability to understand others' beliefs and then infer their intentions in a social context into two components. The first component is the well-established cognitive theory known as "\textbf{Theory of Mind (ToM)}", which is the ability to understand and take into account the mental state of another individual \cite{premack1978does}. The fundamental unit of our theory of mind is our beliefs which are shown in psychological experiments\cite{baron1985does,carpenter2002new}. A belief is an attitude to a content that plays a particular psychological role in us and further directly impacts our actions\cite{Apperly2009DoHH}. The second component is "\textbf{Object-Context Relations}", the perception of any object occurring in some context, and this contextual information provided to us in both implicit and explicit ways can influence how we sense and think of the functionality and utilities of objects \cite{ho1987representing,osiurak2010grasping,bornstein2011perception}. Therefore, this paper aims to uses the combined knowledge of these two components to investigate methods for enhancing collaboration between humans and autonomous systems in environments where verbal communication is prohibited.

We propose BOSS: A Benchmark for Human \underline{B}elief Prediction in
\underline{O}bject-context \underline{S}cenario\underline{s}, a large-scale machine theory of mind video dataset with object-context situations to facilitate research on machine theory of mind in an object-context scenario. In addition to the BOSS dataset, we also developed a novel method for extracting accurate ground-truth labels of mental belief states. The dataset is well-grounded with additional input modalities (e.g. gaze, pose and contextual information). This dataset aims to establish a standard for evaluating novel machine learning models that detect human inner belief states in an object-context setting. To highlight the importance of the various visual modalities in identifying theory of mind and the significance of object-context relations in such contexts, we assessed the dataset using several state-of-the-art deep learning baselines with changing input modalities and masked context objects scenarios.  In addition, we did an in-depth analysis of the results and provided our insights on how different input modalities may have varying effects on the performance of the models and the significance of having a context for forecasting belief states. We aim to establish this dataset as a new and difficult standard for predicting a person's belief states from video, and to extend it as a challenge to the broader vision and robotics community to solve this problem.

Our contributions include:(a) Proposing a novel task of detecting humans' belief in an object-context setting with nonverbal communication. (b) We curated a new large-scale video dataset with multiple input modalities for human belief prediction in object-context scenarios, and introduce a novel approach to obtain precise ground-truth labels for annotating mental belief states. (c) We evaluated the BOSS dataset with various deep learning baselines through a series of comprehensive experiments, and further provide in-depth analysis to how the input modalities and object-context relations impact the learning performance.

\begin{figure*}[ht]
    \centering
    \includegraphics[width=\linewidth]{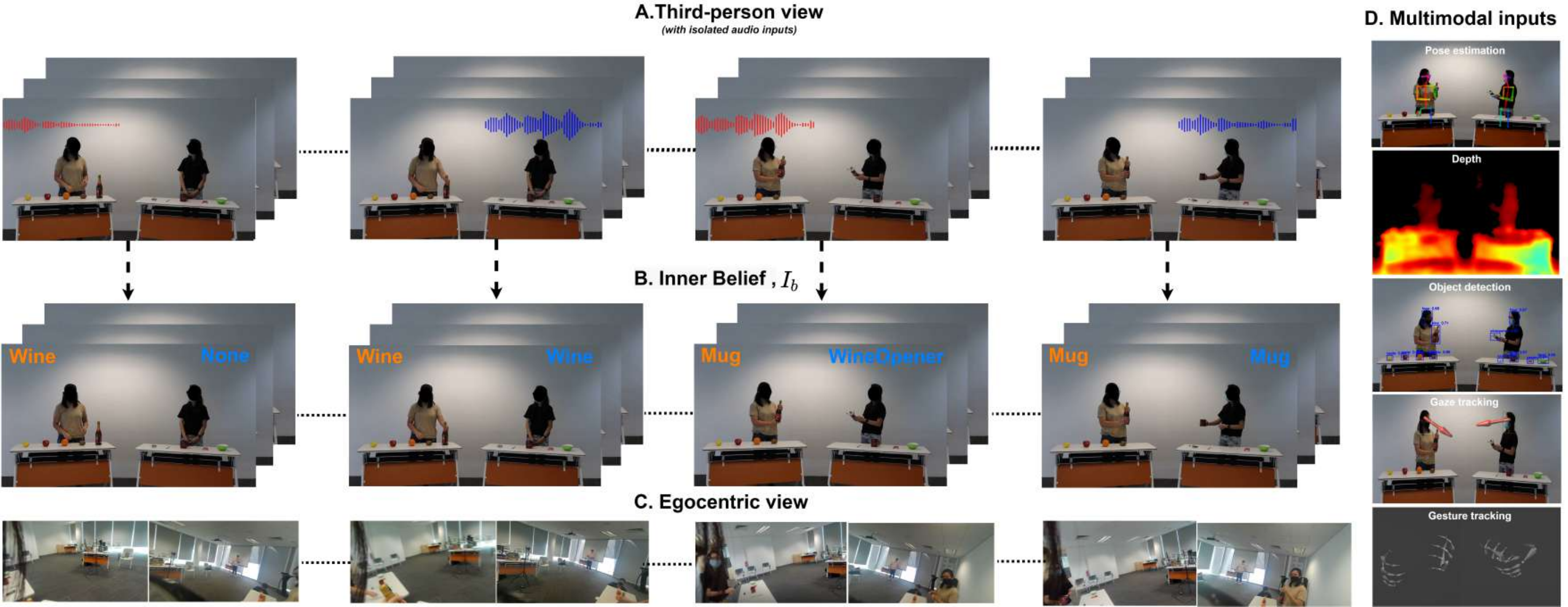}
    
    \caption{Examples from the BOSS dataset: (A) Frames of third-person view of the interactions between the subjects with their individual isolated inner voice. (B) A visualization of the annotated inner belief dynamics across time. (C) The egocentric perception of the subjects. (D) Multimodal inputs: pose, depth, objects, gaze and hand gestures.}
\label{fig:1}
\end{figure*}

\section{Related Work}
\label{related_work}
Due to the difficulty of modeling human beliefs and intentions, related work in scene interpretation and computer vision is limited. In addition, machine theory of mind benchmarks in an object-context scenario are often scarce. Related datasets are typically categorized under the inference of the mental states or beliefs of others using Bayesian inverse planning \cite{ullman2009help,baker2009action}, Partially Observable Markov Decision Processes (POMDP) \cite{doshi2010modeling,baker2011bayesian,de2013much,han2018learning,tejwani2022social}, theory-based modeling of
social goals \cite{baker2008theory,kiley2013mentalistic,baker2014modeling,sukthankar2014plan,kleiman2016coordinate,rabinowitz2018machine} and even as a reinforcement learning problem \cite{wunder2011using,hadfield2016cooperative}. More often than not, a dataset that arises for the study of machine theory of mind focuses primarily on either a 2D grid-world setting \cite{rabinowitz2018machine} or a game built for reinforcement learning \cite{nguyen2020theory,fuchs2021theory}. Therefore, there is a dearth of real-world data of sufficient quality in existing databases.

One dataset recently released by Fan et al. \cite{fan2021learning} is a 3D video dataset featuring social interactions in nonverbal communication events. Their dataset primarily focuses on a triadic environment and attention following, and the proposed objective of predicting belief dynamics is roughly analogous to keyframe video summarization. The BOSS dataset, on the other hand, focuses on capturing the subject's belief states throughout an extended sequence of social interactions in an object-context setting for collaborative work, hence enhancing the dataset's complexity.

\begin{figure*}[ht]
    \centering
    \includegraphics[width=\linewidth]{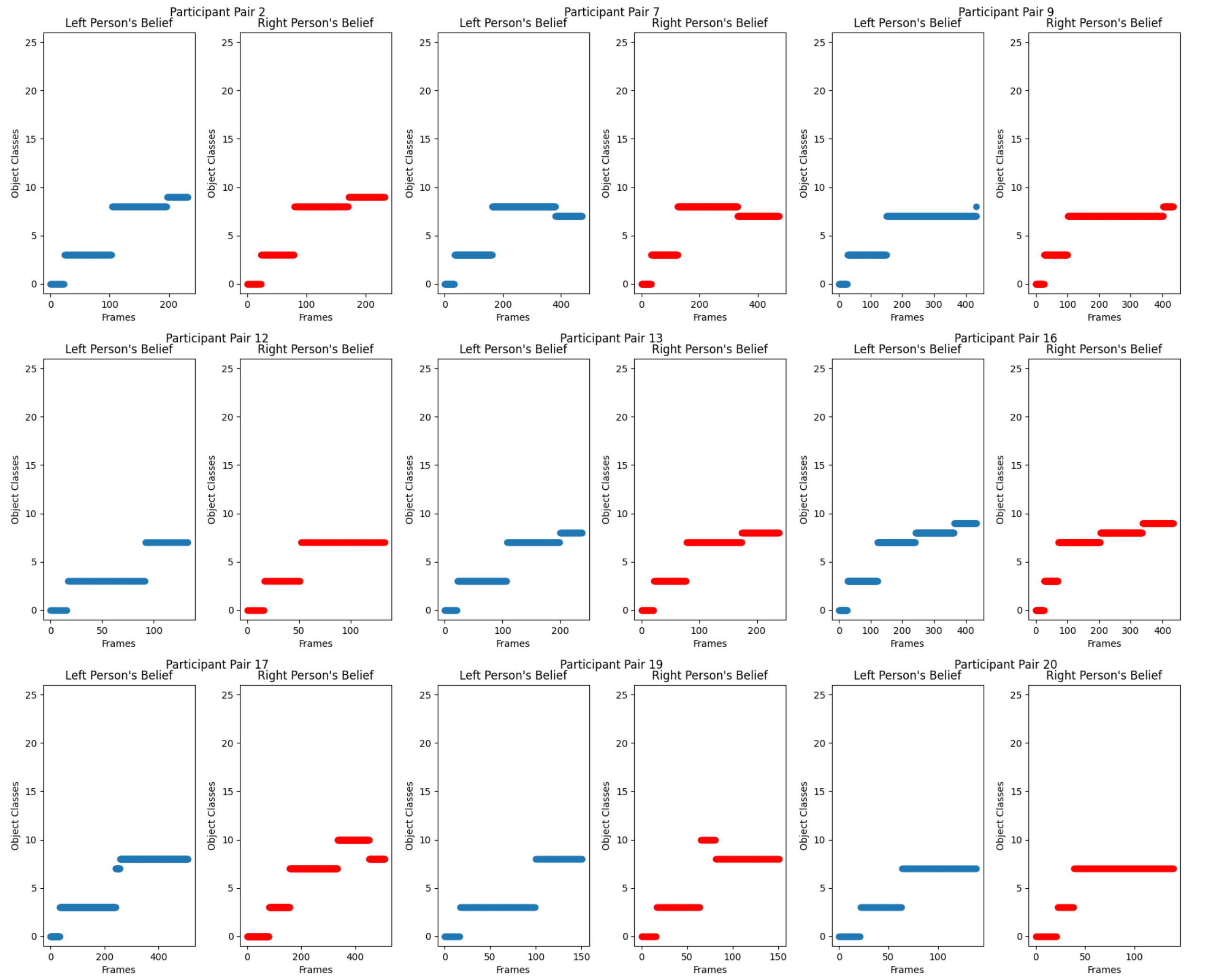}
    
    \caption{Example plots of the belief states (y-axis) across time (x-axis) of the left subject (blue) and right subject (red), for the contextual object class of \emph{lemon} for a few groups of participants. From these few examples, we can clearly observed the difference in sequence lengths, object classes, and delay frames between the belief updates between the two participants.}
\label{fig:3}
\end{figure*}

\section{The BOSS Dataset}
\label{Dataset}
To facilitate research on machine theory of mind in object-context settings, we introduce the BOSS dataset. BOSS is a 3D video dataset compiled from a sequence of social interactions between two individuals in an object-context scenario. The two participants are required to accomplish a collaborative task by inferring and interpreting each other's beliefs through nonverbal communication. Furthermore, we developed a novel method for annotating the individuals' latent mental belief states, for which ground-truth labels are notoriously difficult to obtain.

\subsection{Data Collection}
Ten pairs of participants (five pairs of friends and five pairs of strangers) were recruited in 15 distinct contexts to compile our dataset. We gathered 900 videos from both the egocentric and third-person perspectives, totaling 347,490 frames. As depicted in Figure \ref{fig:0}B, each participant pair is directed to stand in front of a table, one of which contains a list of contextual objects and the other a collection of objects that can be picked based on the context presented. Each contextual object has at least two and no more than three possible combinations of object table selections. The contextual objects, $o_{i}^{context}$= \{\emph{Chips}, \emph{Magazine}, \emph{Chocolate}, \emph{Crackers}, \emph{Sugar}, \emph{Apple}, \emph{Wine}, \emph{Potato}, \emph{Lemon}, \emph{Orange}, \emph{Sardines}, \emph{TomatoCan}, \emph{Walnut}, \emph{Nail}, \emph{Plant}\} and the objects that selected to match these context, $o_{i}^{select}$=\{\emph{WineOpener}, \emph{Knife}, \emph{Mug}, \emph{Peeler}, \emph{Bowl}, \emph{Scissors}, \emph{ChipsCap}, \emph{Marker}, \emph{WaterSpray}, \emph{Hammer}, \emph{CanOpener}\}. All of these objects are widely accessible in households, and the majority of them can be found in the YCB dataset \cite{calli2015ycb}, making it simple to replicate this setup for various robotics tasks. As depicted in Figure \ref{fig:0}B, the individual situated at the contextual objects' table will be given an implicit context task, such as "circle the words on the magazine's cover," and must select the contextual object based on the supplied context. She would then need to communicate her intent non-verbally across the tables to aid the second participant in selecting the correct objects by inferring their intent. We do not supply a script for carrying out the specified activities. We merely present them with the initial context and permit them to communicate their intent non-verbally. To obtain the information necessary to precisely annotate the participants' hidden beliefs in relation to the frame, we requested that all participants wear noise-cancelling headphones and play white noise, and during the data collection process, we asked them to verbalise the names of the items they have in mind whenever their belief is updated. Even though they would not be able to hear one other, this may be done to determine the other individual's intent.

\subsection{Data Annotation}
We also gathered hand gesture data from two Leap Motion sensors placed immediately on the table in front of each participant. The Leap Motion sensors can monitor up to 27 distinct hand elements, such as bones and joints, even when they are obscured by other hand parts. Other input modalities, including object detection, pose estimation, and gaze tracking, were accomplished through a post-processing approach. As illustrated in Figure \ref{fig:1}D, Detecto \cite{Detecto} is used to detect objects, Gaze360 \cite{kellnhofer2019gaze360} is used to collect 3D human gazes, and OpenPose \cite{cao2017realtime} is used to obtain all the critical points for posture estimation. To obtain a precise ground-truth label of their beliefs at each frame, in addition to having an annotator watch the third-person and first-person perspective videos and annotate the subjects' mental belief updates, we use audio clips of the subjects voicing out their inner state to aid the annotator in the annotation process. Despite the fact that it is impossible to achieve a perfect ground-truth annotation due to human reaction time, this method improved ground-truth accuracy because the labels are provided directly by the subjects. After post-processing, the faces of all participants in the movies are obscured for data protection. Therefore, no personally identifiable information is used in training or published to the public.

\begin{figure*}[ht]
    \centering
    \includegraphics[width=\linewidth]{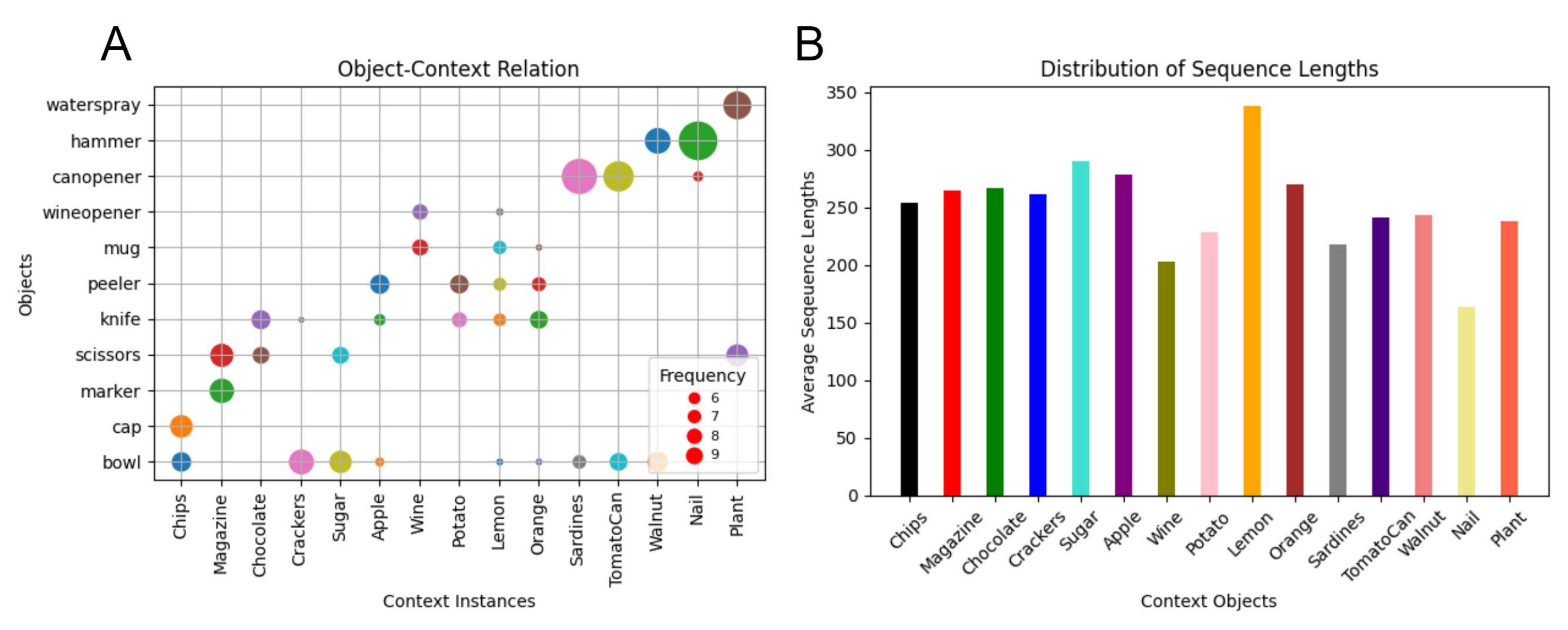}
    
    \caption{Overview of the dataset statistics. (A) The frequency of all the potential object-context matches. (B) The distribution of sequence lengths across the context object classes. }
\label{fig4}
\end{figure*}

\subsection{Dataset Statistics}
The BOSS dataset contains 900 recorded videos captured concurrently from both the egocentric view and the third-person view, yielding a total of 347,490 annotated frames equating to approximately 3.2 hours of video. As depicted in Figure \ref{fig4}B, the average duration of a context object sequence is around 8 seconds, or 250 frames at a frame rate of 30fps. All videos were acquired with audio that was synced. The labeling of belief states in Figure \ref{fig:3} depicts the difference in frames between the initial sign of the mental belief state updates of the object of interest and the beliefs updates of the other participant regarding the same object. As demonstrated in Figures \ref{fig:3} \& \ref{fig4}, the change in belief states across distinct pairs varies throughout the selection object classes. In addition, the dataset was subjected to additional analysis to identify the mean frames of delayed belief updates between the two participants and the distribution of successful inference attempts made by the participants to accomplish the specified task. This extra information is available in the supplementary materials. In addition, Figure \ref{fig4}A demonstrates the matching frequency and probable object-context pairs for all participant pairs. Using the data received from the training set, we can create an Object-Context Relation (OCR) matrix that precisely maps out the interconnected context and object relationships. This matrix is also utilized in training as a type of previous knowledge of the context and object's functionality or utility relationship.

\section{Baseline Evaluation}

     

To evaluate our BOSS dataset, we test various baseline models. In the evaluation, a model needs to select the right beliefs of the two people in each frame. We adopt the average classification accuracy across all beliefs to measure the model performance. We select representative methods for our theory of mind classification task as baselines, which include convolutional neural networks with various common add-ons to capture temporal dependencies.

\textbf{Random} randomly selects a belief for each person at each time step as answer.

\textbf{CNN \cite{he2016deep}} uses a pretrained ResNet with 34 layers to encode each frame as latent features before the features are fed into two separate feedforward networks for multitask classification.

\textbf{CNN+Conv1D} adds a 1D convolutional layer with a kernel size of 5 after the pretrained ResNet to capture short-term temporal dependencies.

\textbf{CNN+GRU \cite{cho2014properties}} adds a gated recurrent unit (GRU) after the pretrained ResNet to capture long-range temporal dependencies.

\textbf{CNN+LSTM \cite{lstm}} adds a LSTM layer after the pretrained ResNet to capture long-range temporal dependencies.

\subsection{Implementation Details}
For all experiments, we use the 300 third-person videos from BOSS. The 300 videos are randomly split it into 60\% for the training set, and 20\% for the validation and test set each.

We train all models using the Adam optimizer \cite{kingma2014adam} for 5 epochs with a fixed
learning rate of 1e-3 and a batch size of 4. All models are implemented using PyTorch. The convolutional neural network (CNN) used in each model is a pretrained ResNet with 34 layers. Each frame is resized to a fixed size of 128 $\times$ 128 and encoded with the ResNet. Other input modalities at the same time step are flattened, encoded with a fully-connected layer and the ReLU activation function and concatenated with the frame's features. These combined features are then passed into an optional add-on layer to capture temporal dependencies (e.g. an LSTM layer or a GRU layer). Finally, the encoded features are passed into two separate fully-connected layers for classification. The add-on layers that are used to capture temporal dependencies are trained from scratch on the BOSS training set. We use cross-entropy loss to train and classification accuracy to evaluate the models. The best checkpoints are evaluated with the validation set, and tested on the BOSS test set. Experiments were run on NVIDIA A100-SXM4 GPUs on Linux servers and the training time is about 1.5 hours for each individual run.




\begin{table*}[ht]
\begin{center}
    \resizebox{\textwidth}{!}{%
    \begin{tabular}{l|ccccc|cc|c}
    \hline
    \multicolumn{9}{c}{Belief State Prediction Accuracy (\%)}\\
    \hline
    Methods & {RGB} & {RGB+OCR} & {RGB+ObjDet} & {RGB+Pose} & {RGB+Gaze}  & {RGB+OCR+ObjDet} & {RGB+Pose+Gaze} & {All} \\
    \hline\hline
    Random & 3.70 & 3.70 & 3.70 & 3.70 & 3.70 & 3.70 & 3.70 & 3.70 \\ 
    CNN \cite{he2016deep}  & 18.06 & 14.24 & 19.30 & 16.15 & 16.22 & 23.26 & 11.84 & 24.23 \\
    CNN+Conv1D & 9.24 & 9.24 & 22.89 & 10.97 & 9.18 & 19.29 & 13.27 & 23.18 \\
    CNN+GRU \cite{cho2014properties} & 16.42 & 10.81 & 20.64 & 13.11 & 13.12 & 21.55 & 14.05 & 15.24 \\ 
    CNN+LSTM \cite{lstm} & 16.86 & 11.91 & 18.47 & 14.40 & 13.29 & 19.80 & 14.60 & 13.99 \\
     
    \hline
    \end{tabular}
    }
\end{center}
\caption{Belief states prediction accuracy across the various deep learning baselines for the different input modalities, object-context relation (OCR), object recognition (ObjDet), pose and gaze information.}
\label{table:final_results}
\end{table*}

\section{Results and Analysis}
In Table \ref{table:final_results}, we present the performance of our baseline models. From the results, we can conclude that BOSS is a challenging task since the prediction accuracies for diverse models that capture spatiotemporal dependencies are consistently low. However, these models still perform significantly above the random baseline, which has a 3.7\% classification accuracy since it has to select a correct answer from 27 options.

\begin{figure}[ht]
    \centering
    \includegraphics[width=\linewidth]{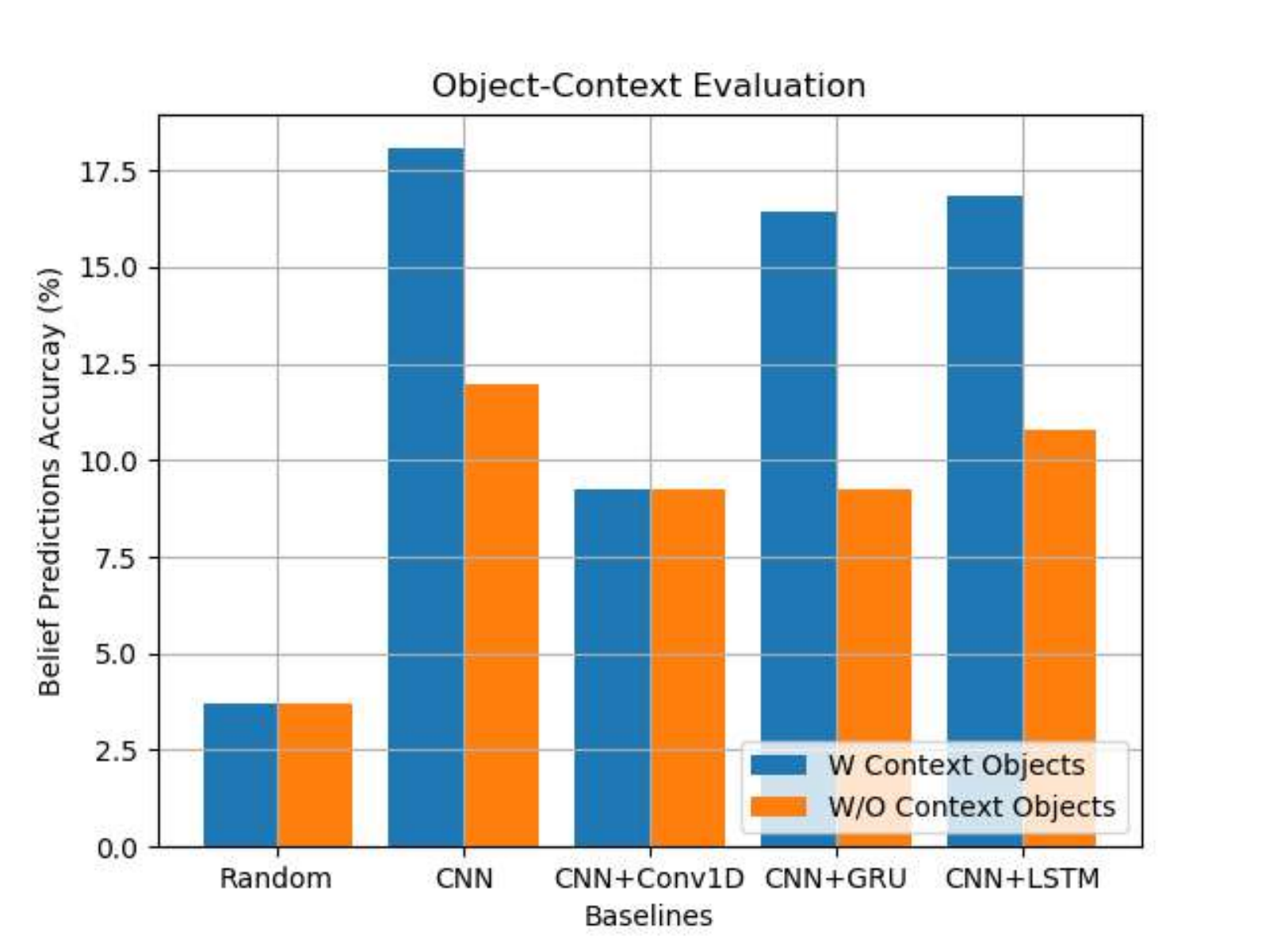}
    
    \caption{Object-context test results. All baselines are trained with RGB inputs and masked contextual objects.}
\label{fig:5}
\end{figure}

\subsection{Model Performance Comparison}
\label{model_performance}
From Table \ref{table:final_results}, we observe that the CNN that is trained with all input modalities have the highest belief state prediction accuracy of 24.23\%, while the CNN+Conv1D has the lowest performance of 9.18\% with RGB+Pose input modalities. However, the CNN+Conv1D has the best performance after the CNN with a prediction accuracy of 23.18\% when it has access to all input modalities. Our results show that models with limited capabilities in capturing temporal dependencies (i.e. CNN and CNN+Conv1D) perform better than models that are trained to capture temporal dependencies of varying length (i.e. CNN+GRU and CNN+LSTM). We hypothesize that this is because the object recognition information significantly reduces the number of possible selections, as shown in Section \ref{modality}, and our CNN+GRU and CNN+LSTM models are not suitable for capturing the temporal dependencies with varying lengths in our long video sequences, such that these temporal dependencies add significant noise instead. This hypothesis is further explored in Section \ref{sequence_length}. Future work that focus on capturing these temporal dependencies better is a promising direction.

\subsection{Input Modality Ablation Study}
\label{modality}
We evaluate the impact of object-context relation understanding, object recognition (ObjDet) and nonverbal communication (Pose/Gaze) information on the performance of our baseline models. The OCR information consists of a fixed 27 $\times$ 27 matrix that maps contextual objects to the selected objects. The OCR information for a contextual object is represented by the probabilities of objects selected for it in the whole training set. The object recognition information is a 27 $\times$ 4 matrix that contains bounding box information for each object in a frame. If an object is present, the information for the object is in the form: $[x_{min}, y_{min}, x_{max}, y_{max}]$. Otherwise, the information will be $[0,0,0,0]$. The object-index mapping for the OCR and object recognition modalities is fixed and identical to the object-index mapping for our ground-truth labels. Pose information is represented by a 25 $\times$ 3 matrix for each person in a frame with 25-keypoint body/hand keypoint estimates. Finally, gaze information is represented by a 3D gaze vector based on the observing camera’s Cartesian eye coordinate system \cite{kellnhofer2019gaze360}.

From Table \ref{table:final_results}, we observe that only the addition of bounding boxes from object detection lead to consistent improvements in performance across all models with an average increase of 5.18\% from the RGB-only models. This is reasonable since knowing the objects present in a frame significantly reduces the number of possible selections. The addition of either OCR, pose or gaze information to RGB inputs generally leads to a decrease in model performance. The addition of OCR causes the most significant decrease in model performance with an average of 3.60\%, compared to a decrease of 1.49\% for human pose information and 2.19\% for gaze information. Since the OCR information is extracted from the training set and static, we hypothesize that our baseline models are overfitting on the additional information. The human pose and gaze input modalities each cause a similar decrease in model performance, possibly because they are added to pretrained ResNet features.

We also run experiments with two additional input modalities to evaluate two distinct categories of information: 1) \textbf{Object-context understanding} and 2) \textbf{Nonverbal human communication signals}. For 1) object-context understanding, we combine OCR and object recognition information with RGB inputs. In general, this combination has the highest average performance increase of 5.83\% from RGB-only inputs across all modality combinations, even though the RGB+OCR setup causes the largest performance decrease. This suggests that object recognition information can help to moderate the effect of the static OCR information. For 2) nonverbal human communication signals, we combine pose and gaze information with RGB inputs. This combination led to an average performance decrease of 1.70\%, which is similar to the performance for RGB+Pose and RGB+Gaze setups.

Lastly, our baseline models obtain an average performance increase of 4.01\% when compared to their RGB-only performance. This is expected considering both the positive and negative effects of the various input modalities.

\begin{table*}[ht]
\begin{center}
    \resizebox{\textwidth}{!}{%
    \begin{tabular}{l|cc|cc|cc}
    \hline
    \multicolumn{7}{c}{Belief State Prediction Accuracy (\%)}\\
    \hline
    Methods & {1 Selection} & {2+ Selections} & {Short Sequences} & {Long Sequences} & {Friends} & {Strangers} \\
    \hline\hline
    CNN \cite{he2016deep} & 19.37 & 16.25 & 17.14 & 19.14 & 21.08 & 15.63 \\
    CNN+Conv1D & 8.65 & 10.04 & 7.67 & 11.10 & 9.96 & 8.66 \\
    CNN+GRU \cite{cho2014properties} & 20.26 & 11.12 & 20.37 & 11.74 & 17.87 & 15.26 \\ 
    CNN+LSTM \cite{lstm} & 14.64 & 19.91 & 12.76 & 21.72 & 15.99 & 17.55 \\
     
    \hline
    \end{tabular}
    }
\end{center}
\caption{Belief states prediction accuracy across the various deep learning baselines for different data categories.}
\label{table:analysis_results}
\end{table*}

\section{Additional Discussions}
We investigate the importance of contextual object presence for to provide more detailed analysis of our proposed BOSS dataset. We also split our test dataset based on 3 metrics: 1) Object selection quantity, 2) Sequence length and 3) Participant relationship type. We evaluate the diverse challenges in our dataset with these data subsets and our baseline models that are trained with only RGB inputs.

\subsection{Object-context Understanding}
To investigate the importance of contextual object presence for training models for accurate belief state predictions, we masked all detected contextual objects from the RGB frames of the video sequences in the training set. This new training set with masked RGB frames is then used to train the various baseline models with only RGB inputs. We evaluate the significance of contextual object presence by comparing the performance of baseline models trained with unmasked sequences and baseline models trained with masked sequences on the test set. The contextual objects in the test set are not masked. Our experiments show that all of the baselines that are trained with the presence of contextual objects in the RGB frames either outperform or are on par with the baseline models that are trained without contextual object presence. The most significant difference in performance is observed with the CNN+GRU with a difference of 7.19\%. The models are improved by an average of 4.83\% when they are trained with contextual objects in their RGB inputs. 

\subsection{Object Selection Quantity}
\label{object_section_quantity}

In our experiments, up to 3 objects are selected by the individual selecting the objects before the desired object is chosen. In the scenarios where 2+ objects are selected, the participant at the table with the contextual objects has to indicate that the object selected is incorrect. This adds diversity to our dataset, and we evaluate the differences in our models' performance between sequences with 1 selection and 2+ selections. There are 40 test sequences with 1 selection, and 20 sequences with 2+ selections. From Table \ref{table:analysis_results}, we can conclude that the CNN+LSTM is significantly better at predicting belief states with RGB-only inputs for sequences with 2+ object selections. This suggests that the LSTM is better at capturing long-range dependencies. There is an average difference of 7.44\% between the CNN+LSTM model and the other models. On the other hand, the CNN and CNN+GRU models are significantly better at predicting belief states for sequences with only 1 object selection.

\subsection{Sequence Length}
\label{sequence_length}
To evaluate how sequence length affects the performance of our baseline models, we split our test set into 2 subsets. We use the median sequence length of our training set to split our test set such that the first subset contains 42 short sequences of length below 240, while the second contains 18 long sequences of length equal to and above 240. From Table \ref{table:analysis_results}, we can see that the results closely mirror the results in Section \ref{object_section_quantity} where the CNN+LSTM performs significantly better on long sequences as compared to the other models. This is reasonable since sequences which contain more than 1 object selection are usually longer. Likewise, the CNN+GRU model performs better on short sequences. The CNN has the least difference in performance between the short and long sequences, which is reasonable since it does not capture temporal dependencies. While the models that attempt to capture temporal dependencies, the CNN+GRU and the CNN+LSTM, outperform the CNN in short and long sequences respectively, their overall performance is worse than the CNN, which does not capture temporal dependencies. Future work that focus on capturing these temporal dependencies across time horizons of varying lengths is a promising direction, as stated in Section \ref{model_performance}.

\subsection{Participant Relationship Type}
The BOSS dataset was collected from 10 pairs of participants, with 5 pairs of friends and 5 pairs of strangers recruited. This is to determine how performance varies between subjects who know and do not know each other. To investigate the relationship between social relations and the accuracy of belief state predictions by deep learning models, we divide the test set into two subsets. The first subset contains 30 sequences with participants who are friends and the other contains 30 sequences with participants who are strangers. The results in Table \ref{table:analysis_results} indicate that the majority of baseline models perform better when testing the sequence of friends versus strangers, with the exception of CNN+LSTM. The difference is especially stark for the CNN with a 5.45\% difference in performance between the two subsets. We can conclude that social relationships have an effect on the models' ability to predict human belief states, especially in cases where temporal dependencies are not captured.

\section{Conclusion}
This paper investigates two crucial components of our social cognitive capacity to identify the belief states of others and accurately infer their intentions in social contexts. Inspired by research in this area of social cognition, we design and propose a new challenging benchmark, BOSS. This benchmark aims to accurately model human belief states in an object-context setting where participants pairs have to communicate non-verbally to complete collaborative tasks. BOSS is one of the largest video datasets with multiple input modalities for predicting human beliefs in object-context settings. The effects of input modalities that relate to nonverbal human communication and object-context relation information on the performance of deep learning baselines are evaluated through extensive experiments. The results of our experiments contribute to a deeper understanding and analysis of one of the most challenging problems: human belief state detection from videos. We believe that our novel dataset on the social aspect of scene understanding based on the prediction of human belief states can positively contribute to robotics, human-robot interaction and other fields.
\newcommand{\squeezeup}{\vspace{-2.5mm}}

\bibliographystyle{IEEEbib}
\bibliography{strings,refs}

\end{document}